\documentclass[runningheads]{llncs}
\usepackage{graphicx}
\usepackage{subfigure}
\begin{document}
\title{TEXT2TASTE: A Versatile Egocentric Vision System for Intelligent Reading Assistance Using Large Language Model}
%
%
\author{
Wiktor Mucha\inst{1} 
\and
Florin Cuconasu\inst{2} 
\and
Naome A. Etori\inst{3}
\and
Valia Kalokyri\inst{4}
\and
Giovanni Trappolini\inst{2}
}
\authorrunning{W. Mucha et al.}
\titlerunning{TEXT2TASTE}
%
\institute{
Computer Vision Lab, TU Wien, Favoritenstr. 9/193-1, 1040 Vienna, Austria \email{wiktor.mucha@tuwien.ac.at} 
\and
Sapienza University of Rome, Rome, Italy
\and
University of Minnesota -Twin Cities, Minneapolis, USA
\and
Foundation for Research and Technology - Hellas, Greece
}
\maketitle              
\begin{abstract}
The ability to read, understand and find important information from written text is a critical skill in our daily lives for our independence, comfort and safety. However, a significant part of our society is affected by partial vision impairment, which leads to discomfort and dependency in daily activities. To address the limitations of this part of society, we propose an intelligent reading assistant based on smart glasses with embedded RGB cameras and a Large Language Model (LLM), whose functionality goes beyond corrective lenses. The video recorded from the egocentric perspective of a person wearing the glasses is processed to localise text information using object detection and optical character recognition methods. The LLM processes the data and allows the user to interact with the text and responds to a given query, thus extending the functionality of corrective lenses with the ability to find and summarize knowledge from the text. To evaluate our method, we create a chat-based application that allows the user to interact with the system. The evaluation is conducted in a real-world setting, such as reading menus in a restaurant, and involves four participants. The results show robust accuracy in text retrieval. The system not only provides accurate meal suggestions but also achieves high user satisfaction, highlighting the potential of smart glasses and LLMs in assisting people with special needs.


\end{abstract}
\section{Introduction}
\label{sec:introduction}

Visual impairment, as one of the aspects of the ageing process, affects a significant proportion of older adults. In 2010, the number of adults aged 50 and over with visual impairment worldwide was estimated to be around 186 million \cite{pascolini2011global}. In the United States alone, the prevalence of uncorrectable vision problems among adults aged 40 years and older exceeded 3 million and is projected to increase to 7 million by 2050 \cite{varma2016visual}. Including those who use corrective lenses, this demographic represents a significant segment of the population and underscores the severity of the problem.
Partial vision loss creates challenges in performing Activities of Daily Living (ADLs) and thus increases older adults' dependence on other people's assistance \cite{owanconversational}. The problem of finding the right detail is exacerbated by the density of text and the volume of information, which corrective lenses cannot solve. The impact of visual impairment on text comprehension extends beyond leisure activities to tasks critical to health maintenance, where access to detailed information, such as reading medication instructions for correct dosage, becomes paramount.
One potential way to address the need to facilitate text comprehension for older adults is through Active Assisted Living (AAL) technologies, in particular wearable egocentric vision devices and algorithms that model behaviours such as action recognition~\cite{mucha2023hands}. These devices can capture images from the user's perspective and then process them to provide augmented reality assistance in the context of recognised actions.
In addition, privacy concerns associated with the use of RGB images~\cite{mucha2022privacy} are minimised, as the egocentric vision system is limited to the task of reading.

Our study proposes using smart glasses with embedded RGB cameras to process textual information in the scene and extract the desired query. Our contribution is listed as follows: 
\begin{itemize}
    \item We implement an egocentric vision system based on the user-friendly Aria \cite{somasundaram2023project} smart glasses with RGB cameras to facilitate text comprehension support. Each video is processed using state-of-the-art object detection algorithms and Optical Character Recognition (OCR) to extract the desired information.

    \item We introduce a system, which combines personal preferences with the digital representation of the text, further processed using the GPT4 \cite{achiam2023gpt} Large Language Model (LLM) to generate a response to the user's query.

    \item We evaluate our method through a specific task of reading menus in a restaurant,
involving four menus from different locations and languages. For this purpose, we
create a demonstration application in the form of a chat that mimics voice control system, which is planned to be used
for real-world applications. The evaluation of our framework with a group
of four persons shows a robust transfer of text into digitised information with 96.77\% accuracy of
retrieval, and accurate meal suggestions from the menu card, with all users being highly satisfied with the performance of the system. 
\end{itemize}


The paper is structured as follows: first, in section \ref{sec:related_work}, we review the existing research on reading assistance systems and identify areas for improvement. In section \ref{sec:methodology}, our methodology is explained in detail. Our findings from evaluation and experimentation are presented in section \ref{sec:evaluation}. Finally, section \ref{sec:conclusion} concludes the study by summarising its main findings.

\section{Related Work}
\label{sec:related_work}
In recent years, several innovative approaches have been proposed to improve the accessibility of written content for visually impaired individuals. These solutions take advantage of technological advances, particularly in image processing, machine learning, and assistive devices. Several studies address the problem using OCR-based methods for translating visual representation into sound. Khete et al. \cite{khete2022autonomous} propose an autonomous assistive system that focuses on converting documents into spoken MP3 data, using OCR and the Google Text-to-Speech API.
Kowshik et al. \cite{kowshik2019assistance} build on the text-to-speech mechanism and extend the functionality by incorporating a pointing finger for control. This interactive approach enhances the user experience, allowing visually impaired people to navigate and interact with digital content in a more dynamic way.
Priya et al. \cite{priya2021assistant} address the limited availability of Braille resources for visually impaired people. In this work, an OCR algorithm processes images and converts them into intermediate text output later transformed into audio using Google Translate.

Our work differs from the presented studies by incorporating comfortable wearable smart glasses and a personalised interaction system. It goes beyond text-to-speech translation by using LLM to create a noise-free digital version of the captured text. 
The user of the device can ask any questions related to the text and receive contextual answers based on personal preferences, exploiting recent advancements in AI \cite{achiam2023gpt,lewis2020rag,trappolini2023multimodal}. Finally, the power of LLM allows performance regardless of the language of the text or the user's query.



\section{Methodology}
\label{sec:methodology}
\begin{figure}[tp]
\centering

\includegraphics[width=\textwidth]{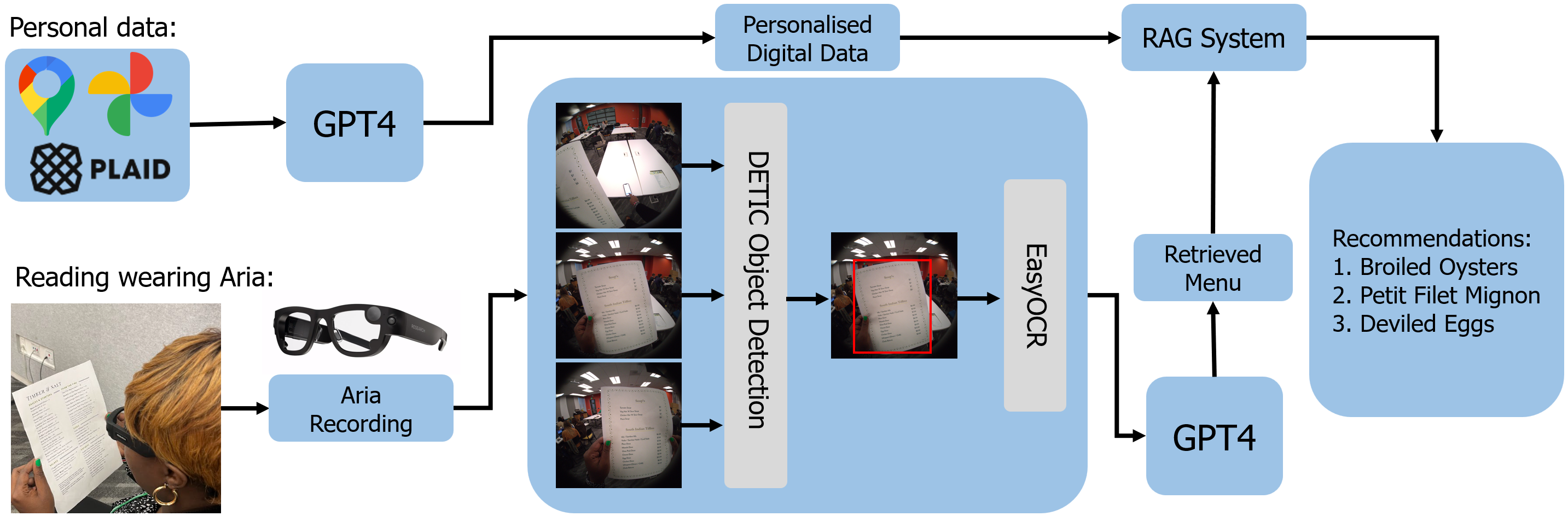}

\caption{Overview of the proposed method for a specific task of reading a menu card. Firstly, the action of reading a menu card is recorded using Aria smart glasses, then this video is processed to select a keyframe and extract the text using EasyOCR and to retrieve the digital menu using the GPT4 model. The digital menu is fed with personalised digital data using the RAG model, resulting in a personalised food preference.}
\label{fig:acc_size}
\end{figure}

\begin{figure}[t]
    \centering
    
    \subfigure[]{
        \includegraphics[height=3cm]{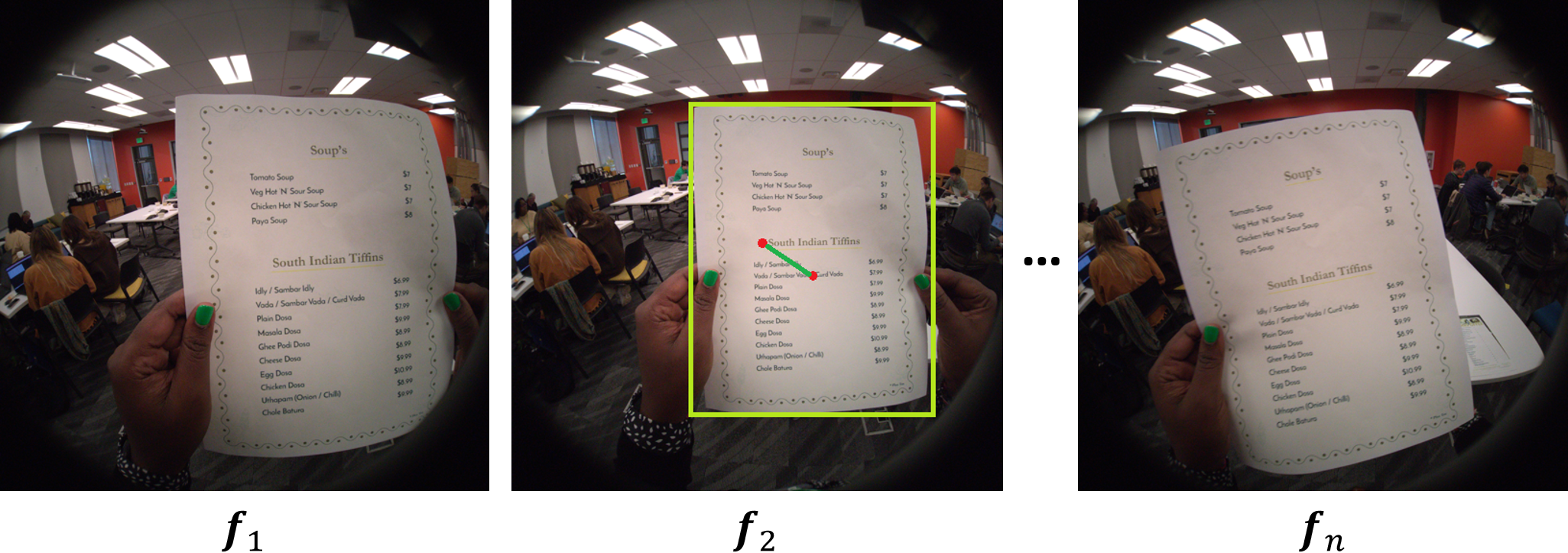}
        \label{fig:frames:subplot1}
    }
    \hfill
    \subfigure[]{
        \includegraphics[height=3cm]{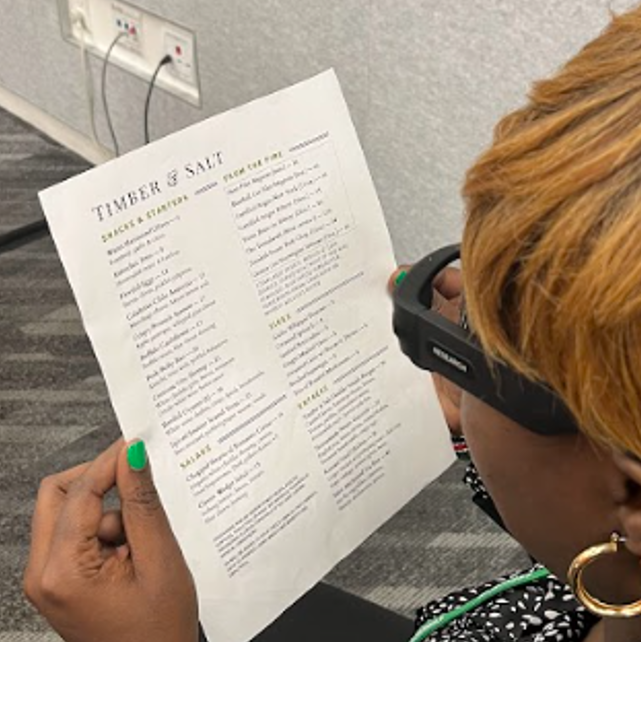}
        \label{fig:frames:subplot2}
    }
    \caption{On the left, frames of a video recorded with Aria. From the sequence of frames, we select the one with the most centred position in the sequence. On the right, a user wearing the Aria device interacts with a menu card.}
    \label{fig:frames}
\end{figure}

We propose a novel text analysis assistant for visually impaired people using Aria smart glasses \cite{somasundaram2023project}. The system uses OCR and LLM to analyse user data extracted from multiple sources and provide contextual guidance. The methodology outlines the experimental design to validate the effectiveness of the system, with a focus on scenarios such as analysing a restaurant menu. The overview of the proposed framework is illustrated in Fig. \ref{fig:acc_size}.

\subsection{Understanding user preferences}

In the setup studied, we focus on food preferences as an example, but the presented system is not limited to this scenario. We extract personal food-related data provided by participants with their consent from bank transaction data via Plaid API, pictures in Google Photos, and favourite restaurants and bars stored in Google Maps. The contextual database is more accurate as the data grows, but privacy is critical in any system, so future real-world solutions should emphasise user security, and this feature itself should only be enabled if the user prefers.

\subsection{Providing text-analysis assistance}

The textual information is extracted from RGB video recordings captured by Aria smart glasses \cite{somasundaram2023project}. User with Aria device reading menu card is shown in Fig. \ref{fig:frames:subplot2}. These egocentric videos serve as the primary data source for our framework. To pinpoint frames of Aria recordings containing relevant menu information, we implement a state-of-the-art per-frame object detection approach, DETIC, following Zhou et al. \cite{zhou2022detecting}. Our algorithm identifies frames where a menu card is visible within the wearer's field of view. To mitigate potential distortions caused by the camera lenses, the frame with the most centrally positioned menu is selected. Once the target frame is identified, OCR is performed using the open-source EasyOCR\footnote{EasyOCR \url{https://github.com/JaidedAI/EasyOCR}} model on the corresponding image. An example of a video sequence with a selected frame is depicted in Fig. \ref{fig:frames:subplot1}. The OCR extracts text from the menu card within a bounding box. This step ensures the isolation of relevant textual content for further processing. To create a digital version of the menu card, we leverage a LLM GPT4\cite{achiam2023gpt}. This model processes the extracted text, transforming it into a structured and readable format. The use of a LLM enhances the accuracy and efficiency of the digitisation process, enabling a seamless transition from the physical to the digital realm for information.

\subsection{User interface}

For demonstration purposes, we create a chat-based application to operate the system, aiming to show the potential use of the device with Gradio \cite{abid2019gradio}. In subsequent improvement steps, we plan to replace it with voice control, allowing users to interact with the smart glasses in a comfortable way. The chatbot is developed using GPT4 \cite{achiam2023gpt}, enhanced by a Retrieval-Augmented Generation (RAG) approach \cite{lewis2020rag}, which operates in two distinct phases. First, a retrieval mechanism is used to extract relevant personal data related to the user's request and the menu provided. This information is then fed into the model to generate contextual and personalised responses. In addition to the personal data, the chatbot also processes the OCR-based restaurant menu and the user's specific query regarding dish recommendations. The user interaction typically begins with queries such as ``What do you recommend from the menu?'', marking the initial engagement with the chat interface. 


\section{Evaluation}
\label{sec:evaluation}
The evaluation of our reading assistance system is carried out using the scenario of reading restaurant menus, with different scenes to assess its real-world performance. The choice is motivated by the ability to evaluate complex text information in different language settings. Each scenario simulates different environmental conditions and user interactions, providing a comprehensive evaluation of the system's adaptability.

Our setup includes four menu cards in different languages, e.g., English, Italian, Polish and Greek, overcoming potential language barriers and ensuring wider usability. In the study, four sighted participants aged between 25 and 35 interact with the menu cards using Aria smart glasses \cite{somasundaram2023project}. Each participant speaks a different language and is provided with menu cards in languages they do not understand, ensuring a comprehensive evaluation of the usability of the developed system across different linguistic backgrounds. In all scenarios, the system demonstrates consistently high accuracy in reproducing digital menu content, reproducing 96.77\% of the items listed in the menus. Each participant in the study is asked to rate their interactions with the system and its contextual suggestions on a scale of 1 to 5. All interactions with our system are rated as highly satisfactory recommendations by the participants with an average rating of 4.87. The contextual system allows the suggestion to be tailored to the user's needs, e.g. excluding allergy products using previously established knowledge of the user's preferences. The chat implementation allows unsatisfactory suggestions to be recreated, incorporating improvements.




\section{Conclusion}
\label{sec:conclusion}
This study implemented a system for assisting people with visual impairments based on Aria smart glasses embedded with RGB cameras. Captured videos were processed using DETIC object detection and the GPT4 LLM to extract information from a given text using contextual personal data and provide guidance through a chat-based user interface. Evaluation using the example of restaurant menus showed high accuracy of text retrieval despite multilingual source cards. All participants in the study confirmed a high level of satisfaction with the solution provided. Our research highlights the effectiveness of integrating LLMs with egocentric vision and identifies innovative ways in which these technologies can be used to enhance AAL solutions and make them more accessible to people with special needs.

\subsubsection{Acknowledgements} 


The authors would like to thank Meta Reality Labs for organising and funding the workshop in which this work was undertaken. In addition, Wiktor Mucha was supported by VisuAAL ITN H2020 (Grant Agreement No. 861091) and the Vienna Science and Technology Fund (Grant Agreement No. ICT20-055). Giovanni Trappolini and Florin Cuconasu were supported by PNRR MUR projects PE0000013-FAIR, SERICS(PE00000014), IR0000013-SoBigData.it. During this study, Florin Cuconasu was enrolled in the Italian National Doctorate in Artificial Intelligence at the Sapienza University of Rome.

\bibliographystyle{splncs04}
\bibliography{my_bib}

\end{document}